\relax
\documentclass[twocolumn]{article} %DO NOT CHANGE THIS
\usepackage{aaai19}  %Required
\usepackage{times}  %Required
\usepackage{helvet}  %Required
\usepackage{courier}  %Required
\usepackage{url}  %Required
\usepackage{graphicx}  %Required
\usepackage{amsmath}
\usepackage{amsfonts}
\usepackage{bm}
\usepackage{subfigure}
\usepackage{booktabs}
\usepackage [autostyle, english = american]{csquotes}
\MakeOuterQuote{"}

\usepackage[colorlinks=true, allcolors=blue]{hyperref}

\usepackage{xspace}
\newcommand{\mname}{\texttt{AWE}\xspace}
\newcommand{\mnamesify}{\texttt{AWEsify}\xspace}

\begin{document}
\title{\mname: Asymmetric Word Embedding for Textual Entailment}
\author{Tengfei Ma$^1$
  \qquad Chiamin Wu$^2$ \qquad Cao Xiao$^1$ \qquad Jimeng Sun$^2$\\
 $^1$ IBM Research \qquad $^2$ Georgia Institute of Technology\\
  \texttt{Tengfei.Ma1@ibm.com}, \texttt{cwu392@gatech.edu}, 
  \texttt{cxiao@us.ibm.com}, \texttt{jsun@cc.gatech.edu}\\
}
\maketitle
\begin{abstract}
Textual entailment is a fundamental task in natural language processing. It refers to the directional relation between text fragments such that the "premise" can infer "hypothesis". In recent years deep learning methods have achieved great success in this task. Many of them have considered the inter-sentence word-word interactions between the premise-hypothesis pairs, however, few of them considered the "asymmetry" of these interactions. Different from paraphrase identification or sentence similarity evaluation, textual entailment is essentially determining a directional (asymmetric) relation between the premise and the hypothesis. In this paper, we propose a simple but effective way to enhance existing textual entailment algorithms by using asymmetric word embeddings. Experimental results on SciTail and SNLI datasets show that the learned asymmetric word embeddings could significantly improve the word-word interaction based textual entailment models. It’s noteworthy that the proposed AWE-DeIsTe model can get 2.1\% accuracy improvement over prior state-of-the-art on SciTail.
\end{abstract}

\section{Introduction}

Textual entailment is an important task in natural language processing (NLP). It refers to the directional relation between text fragments such that the "premise" can infer "hypothesis". To determine such relations requires semantic understanding of text, thus textual entailment is used to measure natural language understanding and can also be applied to other tasks, such as question-answering~\cite{khot2018scitail,chen2017natural}, information retrieval ~\cite{clinchant2006lexical} and document summarization~\cite{pasunuru2017towards}. The objective of textual entailment is to classify a premise-hypothesis (called P-H for simplicity) pair into two classes (entail or non-entail) or three-classes (entail or contradict or neutral). The following is an example from the SNLI dataset~\cite{bowman2015large},  labeled as "entailment", which means the premise implies the hypothesis.
\vspace{-0.1in}
% \begin{quotation}
% \item \noindent \textbf{Premise}: {\it Beats are the periodic and repeating fluctuations  heard  in  the  intensity of a sound when two sound waves of very similar frequencies interfere}.
% \item \noindent \textbf{Hypothesis}: {\it When waves of two different frequencies interfere, beating occurs}.
% \end{quotation} 
\begin{quotation}
\item \noindent \textbf{Premise}: {\it Four guys in wheelchairs on a basketball court two are trying to grab a basketball in midair.}.
\item \noindent \textbf{Hypothesis}: {\it Four guys are playing basketball.}
\end{quotation} 

The approach of textual entailment is generally the same as those of other sentence matching tasks, such as semantic similarity evaluation. However, not all sentence matching tasks are created equal. Different from semantic similarity evaluation, the semantic relation between the premise-hypothesis (P-H) pair is asymmetric in a textual entailment task, i.e. Dis(P,H) is different from Dis(H,P). Consider the word-mover-distance~\cite{Kusner:2015:WED:3045118.3045221} as an example, if we measure the semantic distance of the sentences based on symmetric word embeddings, Dis(P,H) will be equal to Dis(H,P) if we use the same word embedding space for both P and H. However this is not true for textual entailment. For example, for the example P-H pair above, if we exchange the roles of premise and hypothesis, the classification result would be different since the latter sentence cannot entail the former sentence. A symmetric distance between P and H cannot faithfully model the logical relation between texts. Therefore it is better to build an entailment model to retain the asymmetric relations between premise and hypothesis. 
% This idea inspires a series of studies including deep entailment algorithms as reviewed in Related Works.

More recently, word-word interaction attracts much attention especially for the deep learning based textual entailment algorithms~\cite{rocktaschel2015reasoning,parikh2016decomposable,yin2018end}. These algorithms encourage reasoning over entailments of  word- or phrase- pairs. However, the interactions are mostly measured by the symmetric similarity between words, while the directional entailment (i.e., inference) relation is ignored.

In this paper, we propose to utilize the asymmetric word embeddings to improve word-word interactions in textual entailment models. In particular, we first find the candidate entailment word pairs in the training data, and get asymmetric word embeddings for premise sentences and hypothesis sentences separately. The learned embeddings are then used to get entailment based word-level  interaction. They are fed into existing word-word-interaction based textual entailment models (e.g. Decomposable attention model ~\cite{parikh2016decomposable} and DeIsTe ~\cite{yin2018end}). By adding the entailment interaction to the  similarity-based word-word interactions generated by these models, our approach improved these models significantly as demonstrated on SNLI and SciTail data. Especially, we achieved the state-of-the-art performance on SciTail.

To summarize, our work has the following contributions.
\begin{itemize}
\item  We propose a new type of asymmetric word embedding which is motivated by the asymmetric relation of entailment inherent in the texts. Unlike existing asymmetric embedding methods, the proposed approach does not require any external knowledge.
\item  The asymmetric word embedding generated by our approach are general and transferable.  They can be used to improve various word-word interaction based models (as we will show later in DeIsTe\cite{yin2018end} and Decomp-Att\cite{parikh2016decomposable}). Moreover, the extracted entailment word pairs and the derived asymmetric word embeddings can be used for other tasks.  Cross-data test accuracies also demonstrated the comparable performance as training and testing on the same data.
\item By modifying the word-word interactions using asymmetric word embeddings, the  models that use our proposed embeddings (\mname-DeIsTe and \mname-Decomp-Att) gained significant performance improvement over the original models (DeIsTe and Decomp-Att) and state-of-the-art test accuracy on SciTail.
\end{itemize}

\section{Background}
\subsubsection{Word Embedding}
Word embedding is a natural language processing technique that learns continuous vector representations of words, which can be used to represent the semantic similarity of words or support other downstream NLP tasks. These word representations are generally derived from the co-occurrence statistics of a large corpus. For example, the word2vec ~\cite{mikolov2013distributed} is based on the co-occurrence of words in a context window.  %As a result, these context-based word embeddings can be used to predict the context words or determine the semantic similarity. 
%However, they cannot capture the asymmetric word relationship, for example, entailment \cite{vulic2017specialising}, hypernymy\cite{yin2018term}. 

In this section, we will first briefly review one of the most popular word embedding methods: continuous skip-gram ~\cite{mikolov2013distributed}, which is closely related to our later proposed methods. Given a sequence of training words $w_1,w_2,...w_T$, the objective is to maximize the following log-likelihood given by Eq.~\ref{eq:loglike}. 
\begin{equation}
\sum_{t=1}^{T} \sum_{c\in\mathcal{C}_t}\log p(w_c|w_t)
\label{eq:loglike}
\end{equation}
where $\mathcal{C}_t$ denotes the indices of words surrounding word $w_t$. The probability of observing a context word $w_c$ given $w_t$ is given by a softmax function in Eq.~\ref{eq:softmax}.
\begin{equation}
p(w_c|w_t) = \frac{exp({\bf v}_{w_c}^T {\bf u}_{w_t})}{\sum_{w=1}^{W} exp({\bf v}_w^T {\bf u}_{w_t})}
\label{eq:softmax}
\end{equation}
where $u_w$ and $v_w$ are the "input" and "output" vectors of a word $w$. This objective can then be approximately optimized by hierarchical softmax or negative sampling, and we get the vector representations of words.

Word embedding has been a fundamental component for deep learning based textual entailment models. However, most previous works do not differentiate the word embeddings of the premise and the hypothesis, thus the inter-sentence word-word interactions using the word embeddings are usually symmetric. Symmetric similarity-based interactions are not sufficient to represent an entailment relationship. We will introduce our new approach to deriving asymmetric word embeddings for entailment in the next Section.
% For example, the objective of the Skip-gram model is to predict the occurrence of neighboring words, thus the derived word representations could be used to measure the semantic similarity or predict its context word; however similarity alone is not sufficient for determining the entailment relation.

\subsubsection{Decomposable Attention Model (Decomp-Att)}
The decomposable attention model~\cite{parikh2016decomposable} is one of the most important textual entailment models that regard word-word interaction as a key component. It computes a soft alignment matrix and decomposes the entailment task into comparisons of aligned words.
Given two sentences $P = ({\bf p}_1, ..., {\bf p}_{l_p})$ and $H = ({\bf h}_1, ..., {\bf h}_{l_h})$, the model first soft-aligns the elements of $P$ and $H$ using a variant of neural attention:
\begin{eqnarray}
\label{eq:attention}
e_{i,j} = F^\prime({\bf p}_i,{\bf h}_j) = F({\bf p}_i)^T F({\bf h}_j)
\end{eqnarray}
Using these attention weights, we can get the new (softly) alignment vectors
\begin{eqnarray}
&{\bm \beta}_i &:= \sum_{j=1}^{l_h} \frac{\exp(e_{i,j})}{\sum_{k=1}^{l_h} \exp(e_{i,k})} {\bf h}_j \\\nonumber
&{\bm \alpha}_j &:= \sum_{i=1}^{l_p} \frac{\exp(e_{i,j})}{\sum_{k=1}^{l_p} \exp(e_{k,j})} {\bf p}_i\\\nonumber
\end{eqnarray}
where ${\bm \beta}_i$ is the subphrase in $H$ which is aligned to ${\bf p}_i$; while ${\bm \alpha}_j$ is the subphrase in $P$ that is aligned to ${\bf h}_j$. 

Then the aligned subphrases (i.e. ${\bf p}_i$ and ${\bm \beta}_i$, ${\bf h}_j$ and ${\bm \alpha}_j$) are compared using a function $G$ and aggregated.
\begin{eqnarray}
&{\bf v}_1& = \sum_{i} {\bf v}_{1,i} = \sum_{i} G([{\bf p}_i, {\bm \beta}_i])\nonumber\\
&{\bf v}_2& = \sum_{j} {\bf v}_{2,j} = \sum_{j} G([{\bf h}_j, {\bm \alpha}_j])\nonumber
\end{eqnarray}
The final prediction of entailment is based on ${\bf v}_1$ and ${\bf v}_2$: 
$Y = H([{\bf v}_1, {\bf v}_2])$, where $H$ is an feed forward network. 

The Decomp-Att has another optional component, intra-sentence attention, to further enhance the performance. As it is optional and not very related to the motivation of our paper, we omit this component and did not use it in our own models. 

\subsubsection{DeIsTe}
Most recently, ~\cite{yin2018end} proposed a new word-word interaction based model and achieved state-of-the-art performance on the SciTail dataset. Given word-word interactions between the premise-hypothesis pair P-H, the model uses a parameter-dynamic convolution to weight important words in P and H and a position-aware attentive convolution to encode the representation and position information of the aligned word pairs. 

The word-word interaction in DeIsTe is simply the cosine similarity ${\bf I}[i,j] = cosine({\bf p}_i, {\bf h}_j)$ between the words in premise and hypothesis. Given interaction $I$, there are three exploration strategies of these interactions.
\begin{enumerate}
\item importance of ${\bf p}_i$: $a_i = \frac{1.0}{1.0+max({\bf I}[i,:])}$
\item soft best match of ${\bf p}_i$: $H\cdot \text{softmax}({\bf I}[i,:])$
\item hard location of best match of ${\bf p}_i$: $\arg\max {\bf I}[i,:]$
\end{enumerate}

The parameter-dynamic convolution is based on the importance of $p_i$, which measures how important a word is in the convolution encoder for $P$. Specially, for each adjacent trigram $p_{i-1}, p_{i}, p_{i+1}$ in $P$, the parameter-dynamic convolution learns the representation 
\begin{equation}
\label{eq:dyn_attention}
{\bf m}_i = \tanh (a_{i-1} {\bf W}^{-1} {\bf p}_{i-1}, a_{i} {\bf W}^{0} {\bf p}_{i}, a_{i+1} {\bf W}^{+1} {\bf p}_{i+1})
\end{equation}
where the parameters ${\bf W}^{-1}, {\bf W}^{0}, {\bf W}^{+1}$ are shared in all trigrams, and ${\bm \alpha}_{i}$ denotes the importance score of the corresponding ${\bf p}_i$. 

The position-aware convolution uses a similar strategy as the decomposable attention model. It also first soft-aligns ${\bf p}_i$ with all words $h_j$ in the hypothesis $H$, and gets a soft alignment vector representation 
\begin{equation}
{\bf \tilde{p}}_i = \sum_j \frac{\exp({\bf I}[i,j])}{\sum_j \exp({\bf I}[i,j])} {\bf h}_j
\end{equation}
In addition, the word-word interaction is also used to find the hard coding of best matched position. For ${\bf p}_i$, we first find the index of the best matched word in $H$:  $x_i = \arg\max_i ({\bf I}[i,:])$ and then encode the index with an embedding matrix $M$ and transfers it into a continuous vector representation ${\bf z}_i = {\bf M}[x_i]$. 

Concatenating the original ${\bf p}_i$ with the new position embedding $z_i$, we get a new hidden state representation ${\bf c}_i$. Then a position-aware convolution works at position $i$ of $P$ as: ${\bf n}_i = \tanh ({\bf W}[{\bf c}_{i-1}, {\bf c}_i, {\bf c}_{i+1}, {\bf \tilde{p}}_i] + {\bf b})$
Both the parameter-dynamic convolution and the position-aware convolution are stacked with a standard max-pooling layer to get the final representation ${\bf r}_{dyn}$ and  ${\bf r}_{pos}$ respectively. Finally
the concatenation $[{\bf r}_{dyn}, {\bf r}_{pos}]$ is fed to a logistic
regression classifier to get the final prediction.

\section{\mname: Asymmetric Word Embedding for Textual Entailment} 
\label{sec:method}

\begin{figure}[ht]
\centering
\includegraphics[width=\linewidth]{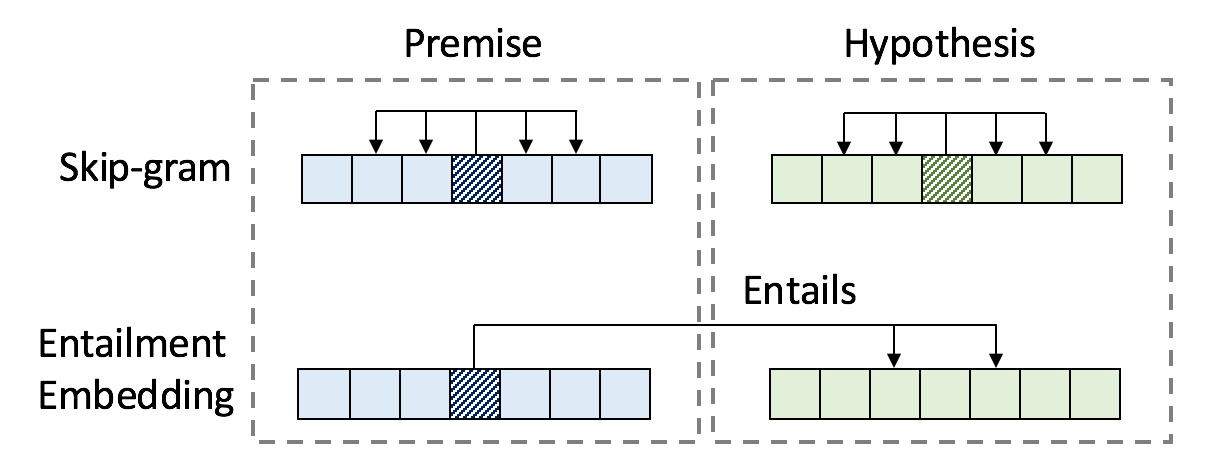}
\caption{Skip-gram Word2vec vs. Entailment Embedding. In the Skip-gram model, we predict whether a word co-occurs with another word in its neighborhood window; while in the Entailment Embedding, given the word in premise, we predict whether it entails another word in hypothesis}
\label{fig:framework}
\end{figure}

Our idea is to model the asymmetric relation of entailment from {\it word level}.  
In a common deep learning approach, the words in the data are from the same embedding space. However, in textual entailment, we would like to model the relationship of $p(w \textbf{ entails } c)$ beyond the relationship of neighborhood co-occurrence that is used in general word embedding techniques ($w$ is a word in premise and $c$ is a word in hypothesis). Thus we modify the Skip-gram word2vec model by selecting different context words. Instead of utilizing the co-occurrence relationships in a neighborhood window, our method creates the entailment contexts as shown in Fig.~\ref{fig:framework}.  

In detail, the proposed approach can be decomposed into the following steps.
\subsection{Entailment Word Pairs}
The most critical step of our method is to create the entailment contexts. Different from previous approaches for asymmetric word embeddings, we do not utilize  existing external knowledge but only focus on the entailment corpus. 

Our assumption is that the entailment sentence consists of multiple word entailment relations. Thus we can find the entailment word pairs from the entailment sentence pairs. Another assumption is that if one word can infer another word, they are generally similar (but not vice versa). So our approach first finds the most similar words in the premise-hypothesis sentence pairs and regard them as possible entailment word pairs.

The aforementioned idea is actually similar to the motivation of using word-word attention in a textual entailment model. However, if the embedding only considers the entailment sentence pairs, it will induce huge amounts of noise.
Let us recall the sentence pair example in the Introduction. By calculating cosine similarities using pre-trained word embeddings, the candidate entailment word-pairs will include ("two", "four") because these two words generally have a high similarity. However, obviously the word pair ("two", "four") cannot represent an entailment relation.  This is to say, using the entailment sentence pairs and word similarity only would bring in some noisy pairs.

One solution to reduce noise in the set of entailment word-pairs is to leverage the neutral (i.e. non-entailment) sentence pairs. If an word pair frequently occurs in both entailment sentence pairs and neutral sentence pairs, it is highly possible that it is noisy. Regarding the previous example, we have another sentence pair marked as "neutral":

\begin{quotation}
\item \textbf{ Premise}: {\it Four guys in wheelchairs on a basketball court two are trying to grab a basketball in midair.	}
\item \textbf{ Hypothesis}: {\it The four players are handicapped.}
\item \textbf{ Label}: neutral
\end{quotation}

The previously mentioned noisy pair ("two", "four") are also included in this neutral sentence pair, so we can possibly remove the pair from the entailment set.

The following is a detailed description of the process to create the set of entailment word pairs:
\begin{itemize}
\item Download pre-trained word embeddings (e.g. pretrained Glove on Wikipedia) and represent each word $w$ as a word vector ${\bf u}_w$.
\item For each entailment sentence pair $P-H$ in the dataset (i.e. labeled as "entailment"), split the sentences into words. For each word $c$ in $P$, calculate the cosine similarity between $c$ and all words in $H$: $cos(c, w) = \frac{{\bf u}_c^T \cdot {\bf u}_w}{||{\bf u}_c|| \cdot ||{\bf u}_w||}$, put all pairs $U_{ent}=\{c,w|cos(w,v)> T_+, w\in P, c\in H\}$ in the entailment word-pair set, where $T_+$ is a similarity threshold.

\item For each neutral sentence pair $S_1-S_2$, also find all word pairs with the similarity larger than a threshold $T_-$: $U_{neu} = \{c,w|cos(c,w)> T_-, w\in S_1, c\in S_2\}$. From the word-pair set $U_{ent}$ remove all word pairs which also occur in the set $U_{neu}$, hence getting the final clean entailment word-pair set $U = U_{ent}/U_{neu}$.

\end{itemize}
In SNLI, there is a third category of sentence pairs labeled as "contradiction", we do not consider this category when extracting the entailment word pairs. Only "entailment" and "neutral" pairs are used.

\subsection{Entailment Word Embedding}

Following the same idea of the Skip-gram model, we train entailment word embeddings on the entailment word-pair set $U$. Denote a word in premise as $c$, a word in hypothesis as $w$, ${\bf v}_c$ as the embedding for a word $c$ in premise and ${\bf u}_w$ for a word $w$ in hypothesis.
The objective is to predict the presence of each entailment word-pair as given by Eq.~\ref{eq:objective}.
\begin{eqnarray}
&&\arg \max_\theta \sum\limits_{c,w}\log p(c|w,\theta) \\\nonumber
=&& \arg \max_\theta \sum\limits_{w} \sum\limits_{c \in N(w)}\log \dfrac{\exp({\bf v}_c^T \cdot {\bf u}_w)} {\sum_{c^\prime}\exp({\bf v}_{c^\prime}^T\cdot {\bf u}_w) }
\label{eq:objective}
\end{eqnarray}
where $N(w)$ is the corresponding word set of $w$ in hypothesis.

This formula is impractical due to the large size of vocabulary (i.e. number of $c^\prime$ in the normalizer). To train the word embedding based on the entailment word pairs, we employ the negative sampling procedure \cite{mikolov2013distributed}. For each word pair $(c,w)$ in $U$, we randomly sample $K$ negative samples for $c$. The original objective $\log p(c|w,\theta)$ is thus approximated by:
\begin{equation}
\log \sigma({\bf v}_c^T\cdot {\bf u}_w) + \sum_{i=0}^K\mathbb{E}_{c_i^\prime \sim P_U(c^\prime)}\log \sigma(-{\bf v}_{c^\prime}^T\cdot {\bf u}_w)
\end{equation}
where $K$ is the number of negative samples for each word $w$; $P_U(c^\prime)$ is the distribution of words in the set $U$; $\sigma$ is the sigmoid function $\sigma (x) = 1/(1+\exp (-x))$. More specifically, if we write the set of negative samples for $w$ as $N_{w,c}$, the objective is rewritten as follows:
\begin{equation}
\label{eq:emb_obj}
\log \sigma({\bf v}_c^T\cdot {\bf u}_w) + \sum_{c^\prime \in N_{w,c}} \log \sigma(-{\bf v}_{c^\prime}^T\cdot {\bf u}_w)
\end{equation}
Notice that different from the original Skip-gram based word embedding which represents a word with two vectors (input and output vectors), in our objective function each word $w$ in the premise only has one vector representation; each word $c$ in the hypothesis also has only one vector representation; and the two vector representations are different for even the same word in the premise and hypothesis. That is to say, the derived word representations are asymmetric (i.e. even for the same word $w$, $u_w \neq v_w$). The asymmetric word embeddings can be directly used for the inference of word-level entailment relationship, e.g. $p(w \rightarrow c) = p(c|w)$ where $\rightarrow$ denotes entailment. Hence, the word-word entailment is  also asymmetric: $p(w \rightarrow c) \neq p(c \rightarrow w)$.

\subsection{Unknown Words in Entailment Word Embedding}
As we can only select a subset of words in the corpus to form the entailment word-pair set, for the remaining words, we will not get the new embedding representations using the aforementioned method. That will result in difficulties in utilizing the new embeddings for textual entailment. To solve the problem of representing the non-entailment words (i.e. the words occurring in the data corpus but not involved in the extracted entailment word-pair set $U$), we bring in two new terms called $UNK_1$ and $UNK_2$ to represent all these words. 

In order to get the embeddings for $UNK_1$ and $UNK_2$, we add them to the negative sample set for each word when we do negative samplings in the algorithm of entailment word embedding. In detail, we change the Equation \ref{eq:emb_obj} into the following form:
\begin{align}
\label{eq:unk_emb_obj}
OBJ = \log \sigma({\bf v}_c^T\cdot {\bf u}_w) + \sum_{c^\prime \in N_{w,c}} \log \sigma(-{\bf v}_{c^\prime}^T\cdot {\bf u}_w) \\\nonumber
+ \log \sigma(-{\bf v}_{UNK_2}^T\cdot {\bf u}_w) + \log \sigma(-{\bf v}_{c}^T\cdot {\bf u}_{UNK_1})\\\nonumber
\end{align}
So if a word in the premise does not occur in the new embeddings, we can use ${\bf u}_{UNK_1}$ to represent it; if a word in the hypothesis does not occur in the new embeddings, we represent it with ${\bf v}_{UNK_2}$.

\section{\mnamesify Textual Entailment Models}
From the process of generating entailment word-pairs, it is easy to see that only a subset of words are contained in the entailment word-pair set thus the derived new asymmetric word embeddings will only cover a fraction of words in the corpus. So if we simply take place of all original word embeddings used in a textual entailment model with our new asymmetric word embeddings, we will get a lot of $UNK$s which do not contain much information. Instead, our idea is to utilize the new asymmetric word embeddings to get more accurate alignment in a word-word interaction based model, such as the Decomposable Attention Model (Decomp-Att)\cite{parikh2016decomposable} and DeIsTe\cite{yin2018end}. In this way, we can integrate the power of original word embeddings for basic input representation of a sentence model and the new asymmetric entailment embeddings for better alignment. In the following sections, we will describe how to utilize our new embeddings to modify Decomp-Att and DeIsTe.

\subsection{\mnamesify Decomp-Att}
The word-word interaction in the Decomp-Att model is based on the soft neural attention in Equation \ref{eq:attention}. 
% We use our new asymmetric embeddings to "correct" the interactions:
% $e_{i,j}^\prime = F({\bf p}_i)^T F({\bf h}_j) + \eta F({\bf u}_i)^T F({\bf v}_j)$
% where ${\bf u}_i$ is the entailment word embedding for word $w_i$ in premise $P$, ${\bf v}_j$ is the entailment word embedding for word $w_j$ in hypothesis $H$.
We use our new asymmetric embeddings to get new alignment vectors:
\begin{eqnarray}
&{\bm \beta}_i^\prime &:= \sum_{j=1}^{l_h} \frac{\exp({\bf v}_j^T\cdot {\bf u}_i)}{\sum_{k=1}^{l_h} \exp({\bf v}_k^T\cdot {\bf u}_i))} {\bf h}_j \\\nonumber
&{\bm \alpha}_j^\prime &:= \sum_{i=1}^{l_p} \frac{\exp({\bf v}_j^T\cdot {\bf u}_i))}{\sum_{k=1}^{l_p} \exp({\bf v}_j^T\cdot {\bf u}_k))} {\bf p}_i\\\nonumber
\end{eqnarray}
where ${\bf u}_i$ is the entailment word embedding for word $w_i$ in premise $P$, ${\bf v}_j$ is the entailment word embedding for word $w_j$ in hypothesis $H$.
Next we combine ${\bm \beta}_i^\prime$ and ${\bm \alpha}_j^\prime$ with ${\bm \beta}_i$ and ${\bm \alpha}_j$, and we add the combined vectors to the model for better prediction.

\begin{eqnarray}
&\hat{{\bm \beta}}_i &= \eta {\bm \beta}_i + (1-\eta) {\bm \beta}_i^\prime \nonumber\\
&\hat{{\bm \alpha}}_j &= \eta {\bm \alpha}_j + (1-\eta) {\bm \alpha}_j^\prime \nonumber\\
&{\bf v}_1^\prime& = \sum_{i} {\bf v}_{1,i}^\prime = \sum_{i} G([{\bf p}_i, \hat{{\bm \beta}}_i])\nonumber\\
&{\bf v}_2^\prime& = \sum_{j} {\bf v}_{2,j}^\prime = \sum_{j} G([{\bf h}_j, \hat{{\bm \alpha}}_j])
\end{eqnarray}
where $\eta$ is a learnable mixture weight. The final prediction of entailment is based on ${\bf v}_1^\prime$ and ${\bf v}_2^\prime$: 
$Y = H([{\bf v}_1^\prime, {\bf v}_2^\prime])$, where $H$ is an feed-forward network. For convenience, we call this model {\bf \mname-Decomp-Att} in the following sections.

\subsection{\mnamesify DeIsTe}
To improve DeIsTe, we still first focus on improving the word-word interaction ${\bf I}[i,j] = cosine({\bf p}_i, {\bf h}_j)$. A straightforward way to improve the model is to utilize the new asymmetric word embeddings to enhance the interaction:
\begin{eqnarray*}
% {\bf I}^\prime[i,j] &=& {\bf I}_0[i,j] + {\bf I}_1[i.j]\\
% &=& cosine({\bf p}_i, {\bf h}_j)+ \eta * cosine({\bf u}_i, {\bf v}_j)\\
{\bf I}^\prime[i,j] &=& \max \{ {\bf I}_0[i,j], {\bf I}_1[i.j] \}\\
&=& \max \{ cosine({\bf p}_i, {\bf h}_j), cosine({\bf u}_i, {\bf v}_j) \}\\
\end{eqnarray*}
where ${\bf I}_0$ is the original word-word interaction in DeIsTe, and the new ${\bf I}_1 = cosine({\bf u}_i, {\bf v}_j)$ indicates a new "entailment" probability.

In DeIsTe, the word-word interaction not only impacts on the alignment, but also helps derive an importance score of each word in $P$ (as in Equation \ref{eq:dyn_attention}). For the new alignment calculation, we can directly use the new interaction matrix ${\bf I}^\prime$ which integrates the word similarity and word entailment. However, we cannot directly use it to take place of the original $\max({\bf I}_0)$ in the calculation of ${\bm \alpha}_i$. For the importance score of a word in the dynamic convolution network on $P$, a word with higher entailment probability should be more important. So we change the importance score $a_i$ for ${\bf p}_i$ into the following form:
\begin{equation}
a_i = \frac{1-\min ({\bf I}_1[i,:])}{1+\max ({\bf I}_0[i:])}
\end{equation}
Compared to the one used in DeIsTe, the additional term $1-\min ({\bf I}_1[i,:])$ is used to punish the words with high entailment probability to infer even the noisy words in hypothesis. Generally an importance entailment word in premise can only infer a few words in hypothesis, so if a word has high inference probability (i.e. ${\bf I}$) with all the words in hypothesis, it might be a common word which does not provide useful information for sentence entailment. For convenience, the new model is called {\bf \mname-DeIsTe} in the following sections.

\section{Experiments}

\subsubsection{Data} We used the following two natural language inference datasets: SNLI\footnote{https://nlp.stanford.edu/projects/snli/} and 
SciTail\footnote{http://data.allenai.org/scitail/}. 
\begin{itemize}
\item \textbf{SNLI} Stanford Natural Language Inference (SNLI~\cite{bowman2015large}) is the largest entailment dataset (570k sentence pairs) created by asking annotators to write statements that would be true (or false) for an image given its caption. The label provided in are "entailment", "neutral", "contradiction" and "-". "-" means the annotators cannot reach consensus with each other, thus generally removed in previous works. In this paper We use the same data split as in ~\cite{bowman2015large}.
\item \textbf{SciTail}~\cite{khot2018scitail} is a more recent textual entailment dataset. It comprises 27K sentence pairs marked as entailment or neutral. It is designed from the end task of answering multiple-choice school-level science questions. With the close relationship with the question-answering task, a substantial performance gain on SciTail can be turned into better QA performance.
\end{itemize}

\subsubsection{Baselines}
We want to show that asymmetric entailment embeddings could benefit word-word interaction based textual entailment models. Therefore, we compare our \mname-Decomp-Att and \mname-DeIsTe with their base models: Decomp-Att\cite{parikh2016decomposable} and DeIsTe\cite{yin2018end}. 
Decomp-Att is one of the most applied advanced textual entailment model, and DeIsTe is the prior state-of-the-art model on SciTail. The results of DeIsTe on SciTail are from \cite{yin2018end}, but on SNLI we use the results from our own running experiments, because the values reported in \cite{yin2018end} is from transfer learning (trained on SciTail and tested on SNLI) while in this paper for fair comparison we want to compare with the model trained on SNLI.
In addition, we added HCRN~\cite{tay2018hermitian} as another baseline, which also considers the asymmetric word matching problem in textual entailment but used a different way. 

\subsubsection{Training and Evaluation}
For both datasets, we first extracted the entailment word-pairs using a predefined threshold ($T_+$ and $T_{-}$). The asymmetric word embeddings are then trained on the word-pair set. Next we get the 300 dimensional embedding representations for all the words in the set as well as two special embedding vectors for $UNK_1$ and $UNK_2$. 

For \mname-DeIsTe, we use the same setting and hyperparameters as DeIsTe in its original paper: batch size 50, no dropout, learning rate 0.01, filter width 3, hidden size 300. All words are initialized by 300D word2vec embeddings~\cite{mikolov2013distributed}. For \mname-Decomp-Att, we also follow the hyperparameter settings of the original Decomp-Att model: 2-layers with hidden size 200, batch size 4, dropout raito 0.2, and learning rate 0.05. All wolds are initialized by 300D GloVe embeddings~\cite{pennington2014glove}. All embeddings remain fixed during training. All the models are trained by Adam~\cite{kingma2014adam}.
Notice that in our proposed \mname-Decomp-Att model, we omitted intra-sentence attention which is an optional component in Decomp-Att. However when comparing with Decomp-Att, we still compare with the best performance achieved by Decomp-Att which includes the intra-sentence attention. All methods are evaluated by accuracies on the development set and test set.

\mname-DeIsTe\footnote{Our codes are available at \url{https://github.gatech.edu/cwu392/AWE-model}} is implemented by theano 1.0.2~\cite{2016arXiv160502688short} based on the codes of DeIsTe\footnote{https://github.com/yinwenpeng/Attentive\_Convolution}, and \mname-Decomp-Att is implemented by Pytorch 0.3.1~\cite{pytorch} and based on the Pytorch version of Decomp-Att\footnote{https://github.com/libowen2121/SNLI-decomposable-attention}. For training models, we used a machine equipped with Intel Xeon E5-2640, 256GB RAM, four Nvidia Titan X and CUDA 8.0.

\subsection{Results}
Table \ref{table:res_SciTail} and Table \ref{table:res_SNLI} present the development accuracies and test accuracies on SciTail and SNLI. Both of the proposed models significantly outperform their base models. On test accuracy, by adding asymmetric word embeddings, \mname-Decomp-Att has a performance gain of 2.0\% over Decomp-Att on SciTail and 0.3\% on SNLI; \mname-DeIsTe outperforms DeIsTe by 2.1\% on SciTail and  1.9\% on SNLI. Especially, to the best of our knowledge, the \mname-Decomp-Att achieved the state-of-the-art performance on SciTail within all approaches without using external resources, see SciTail Leaderboard\footnote{http://data.allenai.org/scitail/leaderboard/}. This demonstrates the effectiveness of our approach to improving word-word interaction based textual entailment models.

In addition, we also compared our model with a recently proposed advanced model, HCRN\cite{tay2018hermitian}, which also considers asymmetric word matchings (HCRN does not publish their results on SNLI, so we only compare with it on SciTail). Different from our approach, HCRN still only uses the original symmetric word embeddings as input, but modifies the interaction function and make it asymmetric. It only applies to its own architecture while our method can be used to adapt most existing word-word interaction based models. By adding asymmetric word embeddings, we can also achieve asymmetric word interactions in the model, and the performance of our \mname-DeIsTe is about 4\% higher than HCRN.
% \vspace{-0.1in}

\begin{table}[ht]
\centering
\caption{Dev and test accuracies (\%) on SciTail}
\label{table:res_SciTail}

\begin{tabular}{l|c|c}
\hline
                 & SciTail (dev)       & SciTail (test)       \\
\hline
 HCRN &    79.4         &   80.0          \\\hline
 Decomp-Att      &    75.4         &   72.3            \\
 \mname-Decomp-Att     & \bf 77.1       &   \bf 74.3                  \\\hline
DeIsTe     &   82.4        &      82.1              \\
\mname-DeIsTe      & \bf  85.1   & \bf 84.2   \\ 
\hline

\end{tabular}
\end{table}

\vspace{-0.1in}

\begin{table}[ht]
\centering
\caption{Dev and test accuracies (\%) on SNLI}
\label{table:res_SNLI}

\begin{tabular}{l|c|c}
\hline
                 & SNLI (dev)       & SNLI (test)       \\
\hline
 Decomp-Att     &     86.4        &   86.8            \\
 \mname-Decomp-Att        &   \bf 87.7   &   \bf 87.1             \\\hline
DeIsTe     &      83.3      &      82.2              \\
\mname-DeIsTe       &\bf 85.1    & \bf   84.1 \\
\hline
\end{tabular}
\end{table}

\begin{table*}[ht]
\centering
\caption{Cross-data evaluation of asymmetric word embeddings. "()" indicates the cross-data results, i.e. models are trained/tested on one dataset but the added asymmetric embeddings are trained from a different dataset. The results of \mname-DeIsTe without "()" are using the asymmetric word embeddings trained from the same dataset.}
\label{table:cross}

\begin{tabular}{l|c|c|c|c}
\hline
                 & SciTail (dev)  & SciTail (test) & SNLI (dev) & SNLI (test) \\
\hline
DeIsTe (baseline)     &    82.4   &   82.1     & 83.3 & 82.2      \\\hline
\mname-DeIsTe + Asymmetric Embeddings on SNLI  &  (84.0)      & (83.6)   & \bf 84.6 & 83.7                \\
\mname-DeIsTe + Asymmetric Embeddings on SciTail  &  {\bf 84.4}   & {\bf 84.2}  & (83.2)   &  ({\bf 84.1})      \\
\hline
\end{tabular}
\end{table*}

\subsubsection{Impact of Thresholds}
\begin{figure}[htb]
\centering
\includegraphics[width=\linewidth]{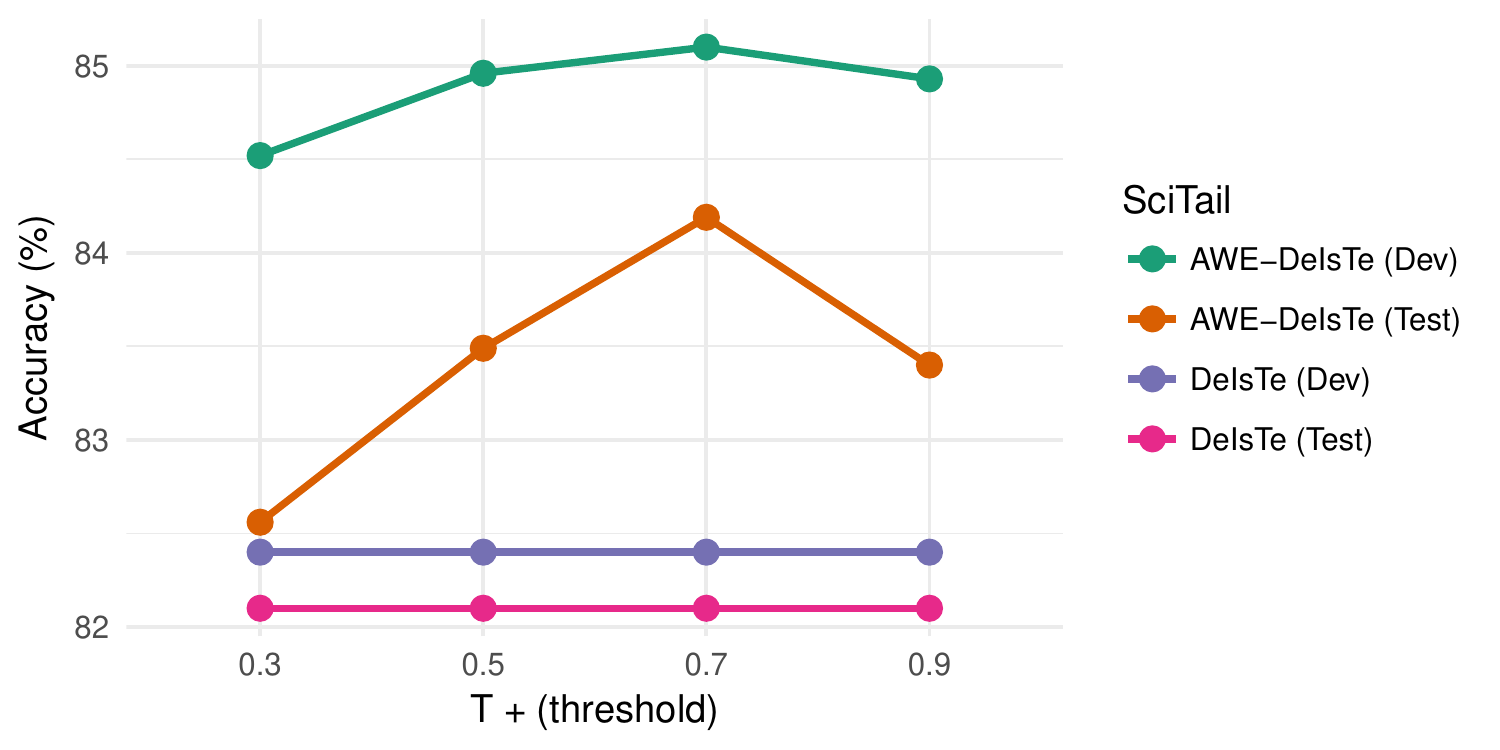}
\caption{Performance Comparison (Accuracy v.s. Entailment threshold ($T_{+}$)) on SciTail. The non-entailment threshold $T_{-}$ is fixed at 0.9.
}
\label{fig:Scitail}
\end{figure}

\begin{figure}[htb]
\centering
\includegraphics[width=\linewidth]{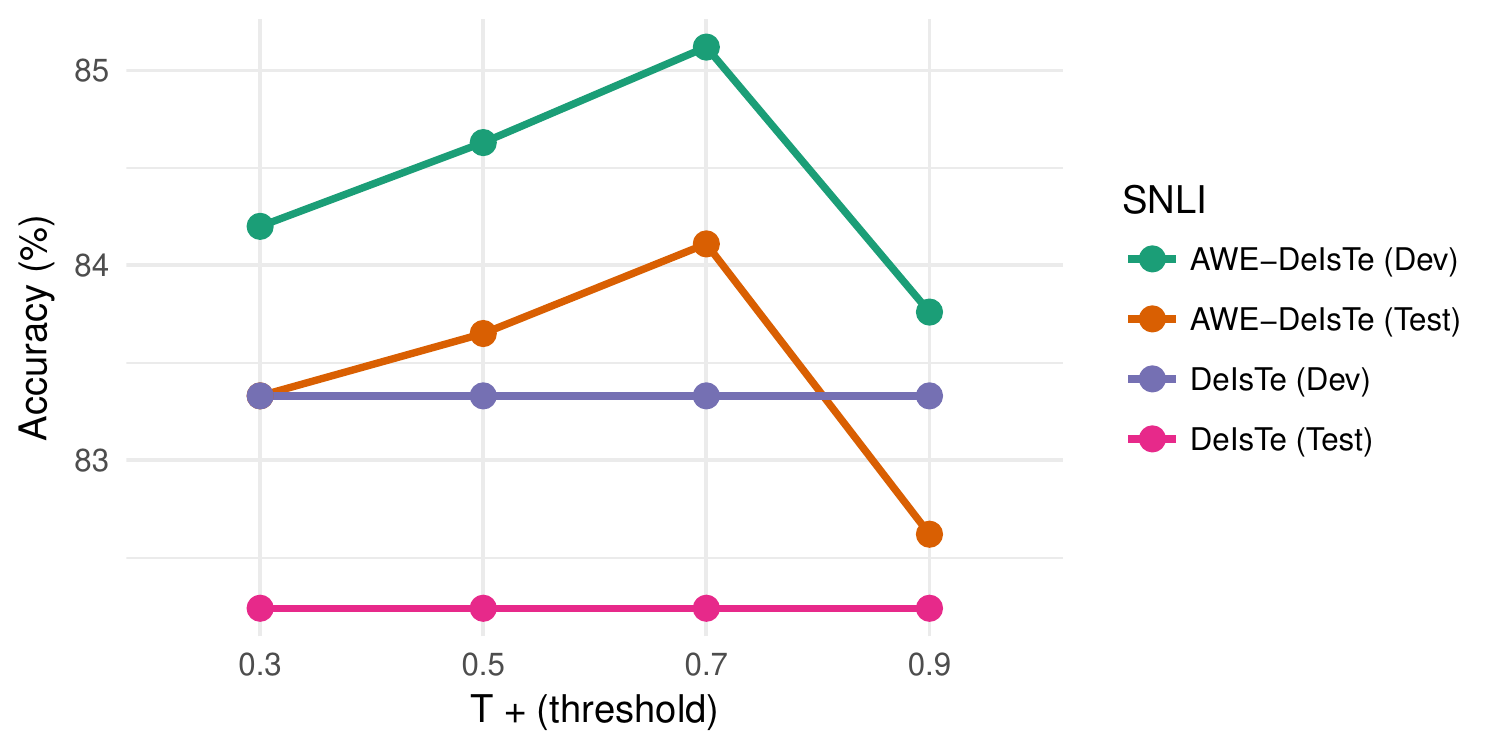}
\caption{Performance Comparison (Accuracy v.s. Entailment threshold ($T_{+}$)) on SNLI. The non-entailment threshold $T_{-}$ is fixed at 0.9.}
\label{fig:snli}
\end{figure}

The thresholds $T_{+}$ and $T_{-}$ are two important parameters to extract the entailment word pairs. Different thresholds will result in different asymmetric word embeddings, thus impacting largely on the textual entailment performance. In Figure \ref{fig:Scitail} and Figure \ref{fig:snli} we show how the performance of DeIsTe changes due to  different thresholds. In order not to bring in too many unrelated word-pairs from the neutral sentence pairs, we fix the non-entailment threshold $T_{-}$ at a high level (0.9) and only change the entailment threshold $T_{+}$. We can see that no matter how we changes the threshold $T_{+}$, the performance (both development accuracies and test accuracies) of \mname-DeIsTe is almost always better than DeIsTe (it does not have a threshold, so the values are unchanged). For both of the datasets, the best accuracies are achieved at $T_{+} = 0.7$ and $T_{-}=0.9$. Then for \mname-Decomp-Att, we fix the threshold and just use the best threshold parameters.

\subsubsection{Evaluation Embedding Transferability}
One advantage of our approach is that the extracted entailment word pairs and the derived asymmetric word embeddings can be further used for other tasks. To demonstrate it, we experimented with \mname-DeIsTe using asymmetric embeddings from other datasets. For example, when we train/test \mname-DeIsTe on SciTail, instead of using asymmetric word embeddings derived from SciTail, we used the alternative asymmetric embeddings from SNLI. Table \ref{table:cross} show the cross-data results of \mname-DeIsTe. Although the new asymmetric word embeddings are not from the same training data, they are still useful to get better results than the baseline model which did not use asymmetric word embeddings. Thanks to the large scale of both SciTail and SNLI, the cross-data test accuracies are even comparable to the original \mname-DeIsTe.

% \begin{figure}[ht]

% 	\begin{subfigure}
%       \centering
%       \includegraphics[width=0.232\textwidth]{ExampleInPaper_Before.png}
%       \label{fig:interaction_sim}
%     \end{subfigure}
%     \begin{subfigure}
%     \centering
%     	\includegraphics[width=0.232\textwidth]{ExampleInPaper_After.png}
%     	\label{fig:interaction_entail}
%     \end{subfigure}
% \caption{Visualization of word-word interactions in an entailment sentence pair. The left is based on word similarity; the right is based on asymmetric word embeddings.}
% \label{fig:visualization}
% \end{figure}

% \subsubsection{Visualization of word-word interactions}
% Visualization of word-word interactions are depicted in Figure \ref{fig:visualization}. In Figure \ref{fig:visualization}, the left figure shows the word-word interaction calculated by word similarity, i.e. the cosine similarity of the original word embeddings. The right figure shows the new word-word interaction calculated by our new asymmetric word embeddings.\js{I don't think we really explain the benefit of our model using this visualization; We should elaborate more and probably move this into supplement if we don't have space}

\section{Related Works}

\subsubsection{Textual Entailment}
Earlier research on textual entailment (somewhere also called natural language inference) were restricted to small data size and relied on hand-crafted features and alignment systems ~\cite{androutsopoulos2010survey}. The development of large-scale annotated datasets ~\cite{bowman2015large} motivated a series of deep neural network based algorithms ~\cite{wang2015learning,chen2016enhancing,liu2016deep} and have led to great progress in textual entailment. They can be categorized as below: 1) sentence encoding based models that find a vector representation for each sentence and classify the concatenation/difference/product of the two vector representations; 2) joint feature models which use cross-sentence features or attentions ~\cite{rocktaschel2015reasoning,wang2017bilateral,parikh2016decomposable}. The sentence matching framework  generally does not differentiate textual entailment with other sentence matching tasks, such as paraphrase identification, sentence similarity evaluation, for example, in ~\cite{kim2018semantic,mccann2017learned}, the same framework can be used for different sentence matching tasks. 

\subsubsection{Word-Word Interaction}
Since the neural attention mechanism has received success in various natural language processing tasks, it is also used in textual entailment for capturing word-by-word interactions ~\cite{rocktaschel2015reasoning,parikh2016decomposable,yin2018end}. In addition, some variants of attention techniques are proposed to improve the interaction architecture, including self-attention~\cite{parikh2016decomposable}, multi-head attention~\cite{vaswani2017attention}, and interactive inference network~\cite{gong2017natural}. Our approach lies on the thread of improving word-word interactions. Unlike previous works, we derive a new type of asymmetric word embedding motivated by the asymmetric relationship of entailments to represent the word-level entailment relation.

\subsubsection{Asymmetric Word Embedding}
Asymmetric word relations have been explored for textual entailment. In hypernymy~\cite{chen2017natural}, the asymmetric word relations were built on external knowledge and were expensive to train. The HRCN~\cite{tay2018hermitian} uses a hermitian co-attention network to model the asymmetric text matching, but it is limited to its own structure. In addition, there are also some works studying the asymmetric word embedding techniques and hypernymy detection\cite{vulic2017specialising,nguyen2017hierarchical}, but these techniques are only concerning about word-level entailment.  
In this work, our proposed approach does not utilize any external knowledge and thus keeps the textual entailment model lightweight.  Moreover, our \mname is a general component which can be added to different word-word interaction based models  as we showed  in DeIsTe\cite{yin2018end} and Decomp-Att\cite{parikh2016decomposable}.

\section{Conclusion}
In this paper we present asymmetric word embedding \mname, a simple but highly effective approach for textual entailment that leverages the asymmetric word-word interactions. We derive a new type of asymmetric word embedding from the entailment word pairs. The new asymmetric word embeddings are then used to adapt existing word-word interaction based textual entailment models, such as Decomp-Att and DeIsTe. The derived new models \mname-Decomp-Att and \mname-DeIsTe sigfinicantly outperforms their base models and \mname-DeIsTe achievs the state-of-the-art performance on SciTail, demonstrating the effectiveness of our approach. Furthermore, we prove that the learned asymmetric word embeddings is even helpful when models are trained on other datasets. Future research includes extending this work to other advanced word-word interaction based models and other asymmetric text matching tasks such as question-answering.

\bibliographystyle{aaai}
\bibliography{aaai}
\end{document}